\documentclass[runningheads]{llncs}
\usepackage[T1]{fontenc}
\usepackage{algorithm}
\usepackage{algpseudocode}
\usepackage{caption} 
\usepackage{color}
\usepackage{hyperref}

\hypersetup{pdflinkmargin=3pt}
\usepackage{xspace}
\usepackage{graphicx}
\usepackage{amssymb}
\usepackage{amsmath}
\usepackage{dsfont}
\usepackage{multirow}
\usepackage{float}
\usepackage{lipsum}
\usepackage{nicefrac}
\usepackage{circledsteps}
\usepackage{enumitem}
\usepackage{colortbl}
\usepackage{graphicx}
\usepackage{booktabs}
\usepackage{siunitx}
\usepackage{breqn}
\usepackage{lipsum}
\usepackage{subcaption}
\usepackage{cleveref}
\usepackage{duckuments}
\usepackage{tikzducks}
\usepackage{tablefootnote}
\usepackage{pifont} 
\usepackage{makecell}
\usepackage{tikz}

\definecolor{lightgray}{gray}{0.95}
\definecolor{midgray}{gray}{0.55}
\definecolor{steelblue}{HTML}{4D82B7}
\definecolor{davysgrey}{rgb}{0.33, 0.33, 0.33}
\definecolor{LightCyan}{rgb}{0.88,1,1}
\definecolor{LightGold}{HTML}{F3E2C5}
\definecolor{ao(english)}{rgb}{0.0, 0.5, 0.0}

\newcommand{\dgreen}[1]{\textcolor{ao(english)}{#1}}

\usepackage[first=0, last=9]{lcg}
\newcommand{\gain}[1]{\dgreen{\plus\textbf{\,#1}}}

\newcommand{\quotationmarks}[1]{``#1''}

\newcommand{\Star}[1]{#1\ensuremath{^*}\kern-\scriptspace}

\newcommand{\tit}[1]{\smallbreak\noindent\textbf{#1 }}

\crefname{section}{Sec.}{Secs.}
\crefname{table}{Tab.}{Tabs.}
\crefname{figure}{Fig.}{Figs.}
\crefname{equation}{Eq.}{Eqs.}
\crefname{algorithm}{Alg.}{Algs.}

\newcommand{\plus}{\texttt{+}}

\newcommand{\er}{ER\xspace}
\newcommand{\der}{DER\xspace}
\newcommand{\derpp}{DER\plus\plus\xspace}
\newcommand{\pp}{\plus\plus\xspace}
\newcommand{\ntu}[1]{Split NTU-#1\xspace}
\newcommand{\methname}{CHARON\xspace}

\makeatletter
\DeclareRobustCommand\onedot{\futurelet\@let@token\@onedot}
\def\@onedot{\ifx\@let@token.\else.\null\fi\xspace}

\def\eg{\emph{e.g}\onedot} 
\def\ie{\emph{i.e}\onedot}

\def\wrt{w.r.t\onedot} 
\def\etal{\emph{et al}\onedot}
\begin{document}
\title{Mask and Compress: Efficient Skeleton-based Action Recognition in Continual Learning}
\titlerunning{\methname}
%
\author{Matteo Mosconi\inst{1}\orcidID{0009-0008-1989-5779} \and
Andriy Sorokin\inst{1}\orcidID{0009-0002-3250-4249} \and
Aniello Panariello\inst{1}\orcidID{0000-0002-1940-7703} \and
Angelo Porrello\inst{1}\orcidID{0000-0002-9022-8484} \and
Jacopo Bonato\inst{2}\orcidID{0000-0001-6751-3407} \and
Marco Cotogni\inst{2}\orcidID{0000-0001-7950-7370} \and
Luigi Sabetta\inst{2}\orcidID{0000-0002-0865-5891} \and
Simone Calderara\inst{1}\orcidID{0000-0001-9056-1538} \and
Rita Cucchiara\inst{1}\orcidID{0000-0002-2239-283X}}
\authorrunning{M. Mosconi \etal}
%
\institute{AImageLab - University of Modena and Reggio Emilia, Modena, Italy\\
\email{name.surname@unimore.it} \and
Leonardo S.p.A.\\
\email{name.surname.ext@leonardo.com}}
\maketitle              
\begin{abstract}
The use of skeletal data allows deep learning models to perform action recognition efficiently and effectively. Herein, we believe that exploring this problem within the context of Continual Learning is crucial. While numerous studies focus on skeleton-based action recognition from a traditional offline perspective, only a handful venture into online approaches. In this respect, we introduce \methname (Continual Human Action Recognition On skeletoNs), which maintains consistent performance while operating within an efficient framework. Through techniques like uniform sampling, interpolation, and a memory-efficient training stage based on masking, we achieve improved recognition accuracy while minimizing computational overhead. Our experiments on \ntu{60} and the proposed \ntu{120} datasets demonstrate that \methname sets a new benchmark in this domain. The code is available at \url{https://github.com/Sperimental3/CHARON}.
\keywords{Continual Learning \and Skeleton Based Action Recognition \and Class Incremental Learning \and Masked Autoencoder.}
\end{abstract}
\section{Introduction}
\textbf{Human Action Recognition (HAR)} has become critical in various domains such as surveillance~\cite{lin2008human,panariello2022consistency}, rehabilitative healthcare~\cite{yin2019skeleton}, and sports analysis~\cite{kong2022human,shao2020finegym}. Early HAR approaches focused on exploiting RGB or gray-scale videos due to their widespread availability. However, recent advancements have explored alternative modalities, including skeletal joints~\cite{du2015skeleton,li2017skeleton,yin2019skeleton}, depth~\cite{sanchez2020exploiting}, point clouds~\cite{fan2021pstnet}, acceleration~\cite{kwapisz2011activity}, and WiFi signals~\cite{sun2022human}. Among these, \textbf{skeleton-based action recognition} stands out as particularly efficient and concise, especially for actions not involving objects or scene context. Skeleton sequences capture the trajectory of key points (\ie, joints) in the human body (\eg, elbows, knees, wrists)~\cite{xin2023transformer}. As joints can be represented by 2D or 3D spatial coordinates, skeletal data offer greater efficiency than images due to the sparsity of skeleton graphs. Moreover, this data structure is robust against changes in appearance, cluttered backgrounds, and occlusion while inherently privacy-preserving~\cite{sun2022human}.

The traditional learning approach to HAR assumes that all necessary data is readily available during training. However, this assumption often does not hold in real-world contexts, as instances or classes may emerge incrementally over time. In such a dynamic context, Deep Neural Networks struggle to acquire new knowledge, often displacing the capabilities acquired during the previous stages. This phenomenon -- widely known as \textit{catastrophic forgetting} -- leads to worse performance and is the focal point of \textbf{Continual Learning (CL)}. Specifically, in the CL setting, models must adapt to address a series of tasks presented sequentially, preserving performance on previously seen ones.

While tasks such as classification~\cite{kirkpatrick2017overcoming,rusu2016progressive,buzzega2020dark,wang2022learning,smith2023coda} and video-based action recognition~\cite{park2021class,castagnolo2023baseline,villa2023pivot} have been widely explored in a Continual Learning setting, skeleton-based HAR has been the subject of limited study in this domain. Although the authors of~\cite{li2021else} have made efforts to address this task, they employ an expandable architecture, which can append a new learnable module to the network each time a new class arises. While such a technique aids in alleviating catastrophic forgetting, the computational footprint of the model gradually grows, making the approach memory-hungry and poorly scalable. Additionally, their setting adds constraints that diverge from real-world scenarios. Namely, they pre-train the network on most training instances and retain only a few classes for the incremental stage.

In this work, we exploit the structure of skeletal data to efficiently store samples in an episodic memory, \ie, a continuously updated \textit{buffer} containing a small subset of past data. Specifically, we enhance the memory efficiency of each sample, thus expanding the effective capacity of the buffer within the same memory allocation. We can do so as skeleton sequences present redundancy in time~\cite{kong2022human}, so they can be compressed by sampling a subset of skeletal poses (\eg, only one every $s$ frames). This operation reduces the temporal resolution of the sequence with minimal information loss. Finally, in later tasks, we reconstruct each retained sample through linear interpolation, which remarkably does not require additional parameters.

We further exploit the redundancy of skeleton sequences by leveraging an approach based on Masked Image Modeling (MIM)~\cite{he2022masked,bao2022beit,tong2022videomae}. Such self-supervised pre-training techniques have recently gained popularity due to the reduced wall-clock time and memory footprint. These methods pre-train a network by feeding it only a portion of the input data and reconstructing it with a lightweight decoder module. Once the pre-training is completed, they discard the decoder and feed the entire input to the model. However, unlike previous works~\cite{wu2023skeletonmae,yan2023skeletonmae}, which employ masking techniques on skeletal data only for pre-training, our approach jointly optimizes both the self-reconstruction and the recognition tasks. Such a choice brings two benefits: \textit{i)} the training time and memory requirements remarkably decrease, and \textit{ii)} the additional reconstruction task acts as a regularizer for the encoder, leading to more meaningful representations.

Finally, at the end of each task, we introduce a \textit{linear probing} phase to better conciliate the self-reconstruction approach with online scenarios. Indeed, if no countermeasures are involved, the encoder may suffer from a covariate shift issue~\cite{ioffe2015batch} during inference, as it has been trained only on a portion of the input but is tested on the whole data. As reported in~\cref{subsec:method}, this may be heavily detrimental to the final classification layer, specifically for high masking ratios. To mitigate such a problem, we freeze the encoder parameters and re-align the classifier in the presence of unmasked input sequences. This process is remarkably lightweight (\ie, optimizing less than 4K parameters for \ntu{60}), yet significantly enhances overall performance.

To assess the proposed approach, we conduct a comprehensive evaluation on the incremental version of two popular datasets, NTU RGB+D 60~\cite{nturose2016dataset} and NTU RGB+D 120~\cite{liu2019ntu}, achieving state-of-the-art performance for class-incremental action recognition in the skeletons domain.

We remark on the following main contributions:
\begin{itemize}[label=$\bullet$]
    \item We reduce the memory requirements of skeleton sequences in the buffer.
    \item We introduce a MIM approach for efficiently handling skeletal data in CL.
    \item We employ a linear probing phase to seamlessly integrate the encoder-decoder approach to the incremental learning setting.
\end{itemize}
\section{Related works}
\tit{Skeleton-based Action Recognition.}~In early skeleton-based action recognition works, sequences were treated as time series, thus processed employing Recurrent Neural Networks (RNNs)~\cite{Du2015HierarchicalRN,zhang2018fusing,hochreiter1997long,chung2014empirical} to capture dynamics over time. These approaches struggled to integrate the spatial context of joints and proved slow and challenging to parallelize. Following works exploited Convolutional Neural Networks (CNNs)~\cite{Kim2017Interpretable3H,ke2017new}, treating skeletal data in various ways to make them compatible with CNNs; some handle coordinates as image channels~\cite{du2015skeleton,li2017skeleton}, while others reshape skeletons by combining joints in space and time~\cite{ke2017new}.

However, these models faced a common limitation: they failed to effectively represent the relationships between skeletal joints moving together in time. Graph Convolutional Networks (GCNs) resolve such shortcomings by exploiting nodes (\ie, joints) temporally and spatially~\cite{yan2018spatial,duan2022dg,duan2022pyskl,chen2021channel,shi2020decoupled}. Subsequently, the emergence of ViT~\cite{dosovitskiy2021an} marked the introduction of transformer-based architectures into computer vision, leading to solutions that integrate self-attention layers into convolutional architectures. One such work, STTFormer~\cite{qiu2022spatio}, divides the sequence in tuples of joints and retains some concepts of CNNs (\ie, pooling aggregation) for in-time features processing. Nonetheless, such an approach under-exploits the sparsity and redundancy of skeletal data. In recent years, masking approaches~\cite{wu2023skeletonmae,yan2023skeletonmae} have been employed to take advantage of these characteristics for pre-training models. In contrast, our proposal adopts the reconstruction objective even during the optimization of the downstream task. Such a choice brings the benefit of reducing the training requirements of the whole pipeline, avoiding the pre-training phase.

\tit{Continual Learning.}~The Continual Learning setting makes a more realistic assumption \wrt standard learning paradigms. Specifically, data arrival is continuous and incremental. A subset of CL is Class-Incremental Learning (Class-IL)~\cite{van2019three}, where the dataset is re-arranged into multiple subsequent tasks, each containing a unique and disjoint set of classes. In this setting, the task identity is not known during inference.

Classical CL methods employ a regularization term that penalizes the alterations of weights to avoid forgetting~\cite{kirkpatrick2017overcoming,zenke2017continual,schwarz2018progress}. Rehearsal methods~\cite{robins1995catastrophic,rebuffi2017icarl,buzzega2020dark,arani2022learning}, on the other hand, employ a limited memory buffer in which they store samples from past tasks and replay them. Another paradigm is represented by dynamic architectures~\cite{rusu2016progressive,bonato2024mind} in which new network components are instantiated for each incoming task; unfortunately, this often leads to a rapid increase in the number of parameters. This approach has been employed by the authors of Else-Net~\cite{li2021else} to tackle skeleton-based HAR in Class-IL. They use the first $50$ classes of NTU RGB+D 60 to pre-train their network, and perform incremental training across $10$ tasks, each focusing on a different class. We retain that such a benchmark diverges from classical CL ones, as it is simplified and far from real-world scenarios. In our work, we utilize the same setting presented by the authors of~\cite{boschini2022class}, who split NTU RGB+D 60 into 6 tasks, each involving \textit{multiple} classes.
\section{Method}
\subsection{Preliminaries}
\label{subsec:preliminaries}
\tit{Class-Incremental Learning.}~In Class-IL, a deep model $f(\cdot;\theta)$ parametrized by $\theta$ is presented with a sequence of tasks $\mathcal{T}_i \text{ with } i \in \{1,\ldots,T\}$, with $T$ denoting the number of tasks. The $i$-th task provides $N_i$ data entries $\{x_i^{(n)},y_i^{(n)}\}_{n=1}^{N_i}$ with $y_i^{(n)} \in \mathcal{Y}_i$; importantly, each task relies on a set of classes disjoint from others s.t. $\mathcal{Y}_i \cap \mathcal{Y}_j = \emptyset \iff i\neq j$. The objective of Class-IL is to minimize the empirical risk over all tasks:
\begin{equation}
    \mathcal{L}_{\text{Class-IL}} = \sum_{i=1}^{T} \mathbb{E}_{(x,y)\sim \mathcal{T}_i} \left[ \mathcal{L}(f(x;\theta), y) \right],
\end{equation}
where $\mathcal{L}$ is the loss function (\eg, the cross entropy for classification) and $y$ is the ground truth label. Since the model observes one task at a time, tailored strategies are required to prevent catastrophic forgetting. Specifically, some rehearsal approaches~\cite{buzzega2020dark,riemer2018learning} employ an additional regularization term $\mathcal{L}_{\mathcal{M}}$ exploiting samples stored in the memory buffer. The objective at the current task $\mathcal{T}_c$ is:
\begin{equation}
    \mathcal{\hat{L}}_{\text{Class-IL}} = \mathbb{E}_{(x,y)\sim \mathcal{T}_c} \left[ \mathcal{L}(f(x;\theta), y) \right] + \mathcal{L}_{\mathcal{M}}.
\end{equation}
\tit{Spatio-Temporal Tuples Transformer (STTFormer).} We adopt as main backbone of our architecture STTFormer~\cite{qiu2022spatio}, a transformer-based model designed for skeleton-based action recognition. It exploits self-attention to capture the cross-joint correlations across adjacent frames. Specifically, a raw skeleton sequence $x \in \mathbb{R}^{C \times F \times V}$, where $C$ is the number of channels (\ie, spatial coordinates), $F$ the number of frames, and $V$ the number of joints, is given as input to the model. This sample is divided into tuples, \ie, sequences of $n$ adjacent frames:
\begin{equation}
    \mathbf{X} = [\mathbf{x}_1,\mathbf{x}_2,\ldots,\mathbf{x}_{\lfloor F/n \rfloor}],\ \text{where}\ \, \mathbf{x}_i \in \mathbb{R}^{C \times n \times V}.
\end{equation}
Each layer of STTFormer comprises two distinct modules, which target either \textit{intra}- or \textit{inter}-tuple relationships. Every element of $\mathbf{X}$ (\ie, each tuple) is first fed to a self-attention layer, which attends the joints in $\mathbf{x}_i$. This phase aims to model the \textit{intra}-tuple characteristics. Then, an \textit{inter}-tuple representation is extracted via temporal pooling.
\begin{figure}[t]
    \centering
    \includegraphics[width=\textwidth]{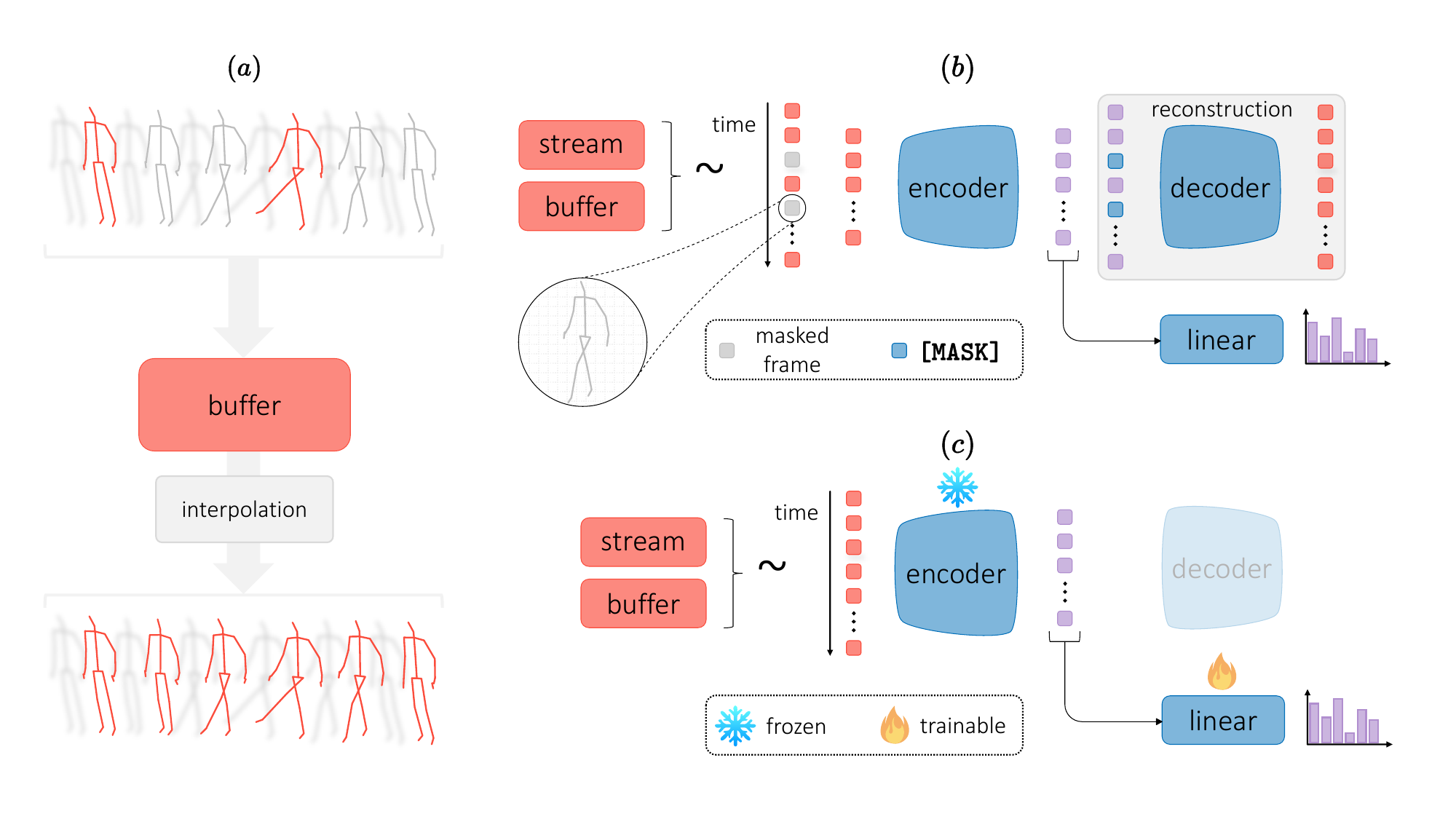}
    \caption{Figure showing the key components of \methname. Our efficient buffer strategy is shown on the left $(a)$. In the upper right $(b)$, we showcase the training phase with the reconstruction regularization, while linear probing is displayed at the bottom $(c)$. Best seen in colors.}
    \label{fig:method}
\end{figure}
\subsection{\methname}
\label{subsec:method}
In this section, we present \methname, which encompasses three components: \textit{i)} a technique to populate the memory buffer, employing linear interpolation to decompress memory samples; \textit{ii)} an efficient training phase with masked inputs; \textit{iii)} a linear probing stage, which refines the classifier and updates the logits stored in the memory buffer. We depict these elements in~\cref{fig:method}.

\tit{Efficient buffer.}~A raw skeleton sequence $x \in \mathbb{R}^{C \times F \times V}$ collects the $C$ coordinates (\eg, $xyz$ in \ntu{60} and \ntu{120}) of $V$ joints at $F$ time instants. Unlike RGB video frames, skeletal data inherently reside in Euclidean space where the concept of distance between points (three-dimensional joints in our case) is well-defined. Additionally, skeleton sequences often exhibit temporal redundancy~\cite{gonzalez2014measuring}. In light of these peculiarities, skeletal data can be easily compressed upon need: for instance, we do so before storing a sequence into the memory buffer. Notably, the compressed sequences can also be reconstructed with minimal loss through a simple linear interpolation. In particular, even with a sampling interval of $s=5$ frames -- \ie, one kept every five, yielding a compression ratio of 80\% -- the reconstructions are close to the raw samples. Based on that, a greater number of instances can be stored within the same memory constraints: in other words, we can accumulate a number of samples $s$ times larger in the buffer.
\begin{figure}[t]
    \centering
    \begin{minipage}{0.44\linewidth}
    \centering
        
    \includegraphics[width=\linewidth]{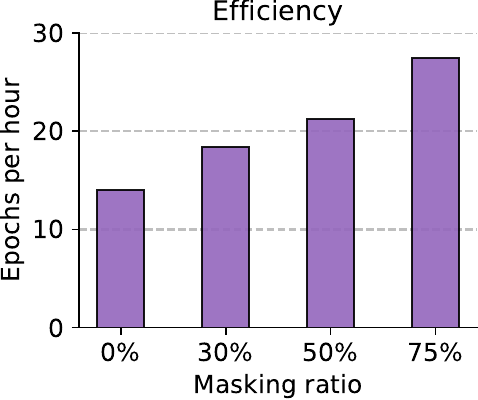}
    \caption{Epochs per hour at different masking ratio values.\protect\footnotemark[1]}
    \label{fig:efficie}
    \end{minipage}
    \hfill
    \begin{minipage}{0.51\linewidth}
    \centering
        
    \includegraphics[width=\linewidth]{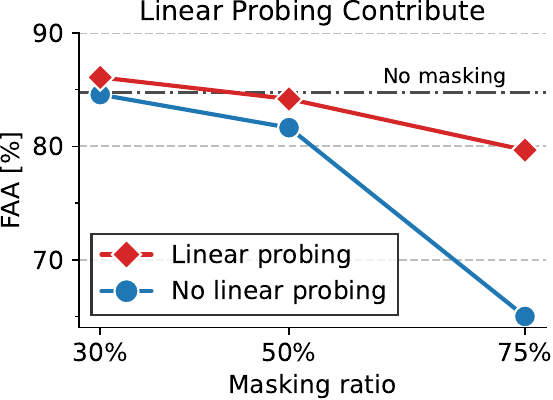}
    \caption{Linear probing contribute on joint training with varying masking ratios.}
    \label{fig:lp}
    \end{minipage}
\end{figure}
\footnotetext[1]{Tests are performed on a single GTX 1080 Ti graphics card.}

When a sample has to be extracted from the buffer for rehearsal, we reconstruct it to obtain $F$ frames again and then treat it as a complete sample. It is noted that, since linear interpolation does not require learnable parameters, the reconstruction of temporal skeletal sequences requires low computational effort.
\tit{Training phase.}~As we mentioned above, a transformer-based architecture founded on~\cite{qiu2022spatio} is adopted as our backbone. We build upon it to derive an encoder-decoder framework inspired by masked autoencoders~\cite{he2022masked}. Notably, this allows us to reduce the computational effort during training, as depicted in~\cref{fig:efficie}. Specifically, given a sample $x$ coming from the current task or the buffer, the first step consists of a linear projection, followed by positional encoding to inject temporal dependencies. Afterward, we feed the encoder $e(\cdot;\theta_e)$ with a temporally masked sample $\tilde{x} \in \mathbb{R}^{C \times \lfloor(1 - \eta) \cdot F\rfloor \times V}$ obtained by dropping a random subset of frames from the input sequence, where $\eta \in [0, 1)$ is the masking ratio.

The encoder projects the input $\tilde{x}$ into the latent space, obtaining features $\tilde{h} = e(\tilde{x};\theta_e)$. From this point, the architecture devises two branches: the first one (\textit{recognition}) features a fully connected layer $f(\cdot;\theta_f)$ to yield pre-softmax logits $z = f(\tilde{h};\theta_f)$. The second branch (\textit{reconstruction}) realizes the self-supervised regularization through a decoder module $d(\cdot;\theta_d)$. Specifically, given the latent feature vector $\tilde{h}$ which has $\lfloor(1 - \eta) \cdot F\rfloor$ tokens, the input of the decoder is formed by filling the missing ones with learnable mask vectors denoted with \texttt{[MASK]}. We place these vectors in the same position as the original masked ones, $h = \textsc{concat}(\tilde{h}, \texttt{[MASK]})$. The training objective is:
\begin{equation}
    \mathcal{L}_{\operatorname{stream}} = \mathcal{L}_{\operatorname{CE}}(z, y)  + \gamma \cdot ||d(h;\theta_d) - x||_2 ^2,
\label{eq:loss_stream}
\end{equation}
where $\gamma$ is a hyper-parameter weighting the impact of the reconstruction loss.

To mitigate forgetting, we incorporate the objective defined in~\cref{eq:loss_stream} into a rehearsal-based framework. Drawing inspiration from~\cite{buzzega2020dark}, we retrieve a mini-batch of samples $x_{\mathcal{M}}$ from the memory buffer at each training step. This mini-batch includes associated predictions $z_{\mathcal{M}}$ (\ie, logits) and labels $y_{\mathcal{M}}$, which are added to the episodic memory along with the corresponding samples. The loss functions for these two components are:

\begin{equation}
    \mathcal{L}_{\operatorname{logits}} = ||f(\tilde{h}_{\mathcal{M}}; \theta_f) - z_{\mathcal{M}}||_2 ^2 \; + \; \gamma \cdot ||d(h_{\mathcal{M}};\theta_d) - x_{\mathcal{M}}||_2 ^2,
\label{eq:loss_logits}
\end{equation}

\begin{equation}
    \mathcal{L}_{\operatorname{labels}} = \mathcal{L}_{\operatorname{CE}}(f(\tilde{h}_{\mathcal{M}};\theta_f), y_{\mathcal{M}}) + \gamma \cdot ||d(h_{\mathcal{M}};\theta_d) - x_{\mathcal{M}}||_2 ^2.
\label{eq:loss_labels}
\end{equation}

\noindent The mini-batch of samples $x_{\mathcal{M}}$ undergoes the same pipeline of the input stream $x$, producing the latent features $\tilde{h}_{\mathcal{M}} = e(\tilde{x}_{\mathcal{M}}; \theta_e)$ and $h_{\mathcal{M}} = \textsc{concat}(\tilde{h}_{\mathcal{M}}, \texttt{[MASK]})$.

The final objective of this phase is:
\begin{equation}
    \mathcal{L} = \mathcal{L}_{\operatorname{stream}} + \alpha \cdot \mathcal{L}_{\operatorname{logits}} + \beta \cdot \mathcal{L}_{\operatorname{labels}},
\label{eq:loss_train}
\end{equation}
where $\alpha$ and $\beta$ are two balancing hyperparameters.
\vspace{0.13cm}
\tit{Linear probing.}~As described above, the model is trained with partial skeleton sequences. While providing an efficient training strategy, there is a factor that could hinder the overall performance during evaluation. Indeed, we argue that the classification heads $f(\cdot;\theta_f)$ could be subject to possible misalignment due to the different conditions we have at training (masking \textit{on}) and test time (masking \textit{off}). To address this issue, highlighted in~\cref{fig:lp}, we devise an auxiliary linear probing stage at the end of each task, which lasts for a few epochs (\ie, 10\% of the number employed for the main training stage). During this phase, only the parameters of the classifier are allowed to change, while the encoder remains frozen. In doing so, we feed each full (\ie, not masked) sample $x \in \mathbb{R}^{C \times F \times V}$ to the encoder.

In formal terms, as for the main training phase, the encoder projects the input $x$ into the latent space obtaining hidden features $h = e(x;\theta_e)$. The fully connected linear layer $f(\cdot;\theta_f)$ produces then the logits $z = f(h;\theta_f)$ to which a cross-entropy loss is finally applied. In this phase, we still employ the regularization from~\cite{buzzega2020dark}. Thus, the resulting objective $\mathcal{L}_{\operatorname{lp}}$ can be written as:
\begin{equation}
    \mathcal{L}_{\operatorname{lp}} = \mathcal{L}_{\operatorname{CE}}(z, y) + \alpha \cdot ||f(h_{\mathcal{M}};\theta_f) - z_{\mathcal{M}}||_2 ^2 \; + \; \beta \cdot \mathcal{L}_{\operatorname{CE}}(f(h_{\mathcal{M}};\theta_f), y_{\mathcal{M}}).
\label{eq:loss_linprob}
\end{equation}
Traditional works using masked autoencoders~\cite{he2022masked,tong2022videomae} typically distinguish between a pre-train phase and one of linear probing to adapt to downstream tasks. However, we argue that leading these stages separately can result in a more cumbersome approach, potentially undermining the efficiency we seek. To solve this,~\cref{eq:loss_train,eq:loss_linprob} are computed sequentially during each task, according to the incremental setting (\ie, holding only a partial amount of data, the one belonging to the current task).
The complete algorithmic procedure for a single task is described in~\cref{alg:derppmae}.
\begin{algorithm}[t]
\caption{Training \methname at the current task}
\label{alg:derppmae}
\begin{algorithmic}
    \State \textbf{Requires}: dataset $D_{\mathcal{T}_c}$, parameters $\theta$ ($\theta_e, \theta_f, \theta_d$), scalars $\alpha$, $\beta$ and $\gamma$, learning rate $\lambda$, masking ratio $\eta$, buffer $\mathcal{M}$. 

    \State \textbf{Main training phase}:
    
    \For{$(x, y)$ \textbf{in} $D_{\mathcal{T}_c}$}
    \State $(x_{\mathcal{M}}, y_{\mathcal{M}}, z_{\mathcal{M}}) \gets \operatorname{interpolate}(\operatorname{extract}(\mathcal{M}))$
    \State $\tilde{x}, \tilde{x}_{\mathcal{M}} \gets \operatorname{random\_masking}(x, \eta), \operatorname{random\_masking}(x_{\mathcal{M}}, \eta)$
    
    \State $\mathcal{L} \gets $ \cref{eq:loss_train,eq:loss_stream,eq:loss_logits,eq:loss_labels}
    \State $\theta \gets \theta - \lambda \cdot \nabla_\theta \mathcal{L}$
    \EndFor
    
    \State \textbf{Linear probing}:
    
    \For{$(x, y)$ \textbf{in} $D_{\mathcal{T}_c}$}
    \State $(x_{\mathcal{M}}, y_{\mathcal{M}}, z_{\mathcal{M}}) \gets \operatorname{interpolate}(\operatorname{extract}(\mathcal{M}))$
    
    \State $\mathcal{L} \gets $ \cref{eq:loss_linprob}
    \State $\theta_f \gets \theta_f - \lambda \cdot \nabla_{\theta_f}\mathcal{L}$
    
    \State $\mathcal{M} \gets \operatorname{populate}(\mathcal{M}, (\operatorname{uniform\_sampling}(x), z, y))$
    \EndFor

\end{algorithmic}
\end{algorithm}
\section{Experimental analysis}
\subsection{Datasets}
\tit{\ntu{60} and \ntu{120}.} NTU is one of the most popular benchmarks for action recognition on skeletal data. Initially comprising 60 classes and \num{56578} samples in its original version~\cite{nturose2016dataset}, and later expanded to 120 classes and \num{113945} samples~\cite{liu2019ntu}, this dataset encompasses a diverse range of actions involving up to two individuals. The data collection process involves three Kinect cameras~\cite{zhang2012microsoft}, positioned with different angles \wrt the subject. They provide RGB videos, IR videos, depth map sequences, and 3D skeletal data. Participants of various ages have contributed to the datasets construction, ensuring its broad applicability and relevance.

We adopt the extraction process employed by~\cite{qiu2022spatio}. As original raw sequences contain a varying number of frames, we apply bilinear interpolation to obtain fixed-length sequences $x$ (\ie, 120 frames) s.t. $x \in \mathbb{R}^{(C=3)\times (F=120) \times (V=25) \times (B=2)}$. The axis identified by $B$ regards the poses of the potentially two subjects involved in the action.

To test our approach in the CL scenario, we embrace \ntu{60}, introduced in~\cite{boschini2022class}, an incremental learning benchmark derived from the standard NTU dataset. The authors of~\cite{boschini2022class} divide NTU RGB+D data into 6 tasks, each defining a 10-class classification problem. We also introduce \ntu{120}, an extension of the previous benchmark. Such a version brings a significant additional challenge, as seen in recent offline literature~\cite{duan2022revisiting,duan2022dg}, leaving the way open for future works in the continual domain. To be as compliant as possible with the previous literature, we keep the original $6$ tasks split and add another $6$, each consisting of $10$ classes, resulting in a $12$ tasks incremental scenario. We describe in the supplementary material the exact order in which classes are split into tasks.

We report results for the cross-subject (XSub) and cross-view (XView) data modalities~\cite{nturose2016dataset} for \ntu{60}, and cross-subject (XSub) and cross-setup (XSet)~\cite{liu2019ntu} for \ntu{120}. 
\subsection{Implementation details}
The custom version we adopt for STTFormer~\cite{qiu2022spatio} reduces the width of intermediate layers to obtain a more lightweight model. We set the number of frames in each tuple $n=6$ as in the original paper. Following the asymmetric design proposed in~\cite{he2022masked}, we employ $8$ layers for the encoder and $3$ for the decoder. We refer the reader to the supplementary material for further details.
Additionally, we employ an $\alpha$ of $0.3$ and a $\beta$ of $0.8$ for~\cref{eq:loss_train,eq:loss_linprob}, while we use a $\gamma$ of $0.5$ in approaches using the reconstruction regularization (\cref{eq:loss_stream,eq:loss_logits,eq:loss_labels}). We adopt a batch size of $16$ for all our experiments with a vanilla SGD optimizer and a learning rate of $0.05$. Each task of the incremental setting lasts for $30$ epochs. With the same hyperparameters as above, we perform $3$ epochs for the linear probing phase. Finally, concerning data augmentation, we follow the original STTFormer implementation, applying a simple random rotation to each input sample.
\subsection{Results}
\begin{table}[t]
\centering

\rowcolors{6}{lightgray}{}
\setlength{\tabcolsep}{3pt}
\renewcommand{\arraystretch}{1.1}

\caption{FAA (\%) results on \ntu{60} and \ntu{120}. For \textbf{\methname}, we report the results with a masking ratio equal to $30\%$. We highlight in green the gains achieved by our approach \wrt the best-competing method.}
\resizebox{\textwidth}{!}{
\begin{tabular}{lcccccccc}
\toprule
 & \multicolumn{4}{c}{\textbf{Split NTU-60}} & \multicolumn{4}{c}{\textbf{Split NTU-120}} \\

\midrule

\textbf{Method} & \multicolumn{2}{c}{\textbf{XView}} & \multicolumn{2}{c}{\textbf{XSub}} & \multicolumn{2}{c}{\textbf{XSet}} & \multicolumn{2}{c}{\textbf{XSub}} \\

\specialrule{\lightrulewidth}{0pt}{0pt}

\textbf{FT} & \multicolumn{2}{c}{16.05\tiny{$\pm$0.07}} & \multicolumn{2}{c}{15.64\tiny{$\pm$0.05}} & \multicolumn{2}{c}{7.19\tiny{$\pm$0.06}} & \multicolumn{2}{c}{6.97\tiny{$\pm$0.23}}\\
\rowcolor{lightgray}
\textbf{JT} & \multicolumn{2}{c}{84.75\tiny{$\pm$0.02}} & \multicolumn{2}{c}{77.32\tiny{$\pm$0.54}} & \multicolumn{2}{c}{71.18\tiny{$\pm$1.07}} & \multicolumn{2}{c}{70.15\tiny{$\pm$0.98}}\\

\specialrule{\lightrulewidth}{0pt}{0pt}

$\mathcal{M}_{size}$ & 500 & 2000 & 500 & 2000 & 500 & 2000 & 500 & 2000\\

\specialrule{\lightrulewidth}{0pt}{0pt}

\textbf{iCaRL} & 51.54\tiny{$\pm$1.3} & 53.41\tiny{$\pm$1.1} & 47.12\tiny{$\pm$1.4} & 50.69\tiny{$\pm$1.2} & 32.91\tiny{$\pm$0.9} & 34.74\tiny{$\pm$0.7} & 33.03\tiny{$\pm$1.3} & 36.68\tiny{$\pm$1.0}\\
\textbf{Else-Net} & 40.81\tiny{$\pm$0.8} & 59.10\tiny{$\pm$0.2} & 39.72\tiny{$\pm$0.4} & 57.00\tiny{$\pm$1.0} & 19.37\tiny{$\pm$0.6} & 33.52\tiny{$\pm$0.6} & 18.43\tiny{$\pm$0.7} & 33.95\tiny{$\pm$0.3}\\
\textbf{\er} & 51.00\tiny{$\pm$1.6} & 68.27\tiny{$\pm$0.1} & 45.80\tiny{$\pm$0.5} & 62.74\tiny{$\pm$1.9} & 26.35\tiny{$\pm$1.1} & 43.12\tiny{$\pm$0.4} & 26.19\tiny{$\pm$1.7} & 45.06\tiny{$\pm$0.7}\\
\textbf{\der} & 51.36\tiny{$\pm$0.9} & 66.74\tiny{$\pm$0.1} & 49.97\tiny{$\pm$1.9} & 63.48\tiny{$\pm$1.3} & 27.83\tiny{$\pm$1.7} & 40.19\tiny{$\pm$0.9} & 30.10\tiny{$\pm$1.5} & 36.10\tiny{$\pm$1.8}\\
\textbf{\derpp} & 60.41\tiny{$\pm$0.5} & 73.09\tiny{$\pm$1.3} & 57.22\tiny{$\pm$1.0} & 67.64\tiny{$\pm$1.6} & 34.27\tiny{$\pm$1.4} & 50.06\tiny{$\pm$0.6} & 36.29\tiny{$\pm$0.3} & 49.81\tiny{$\pm$0.8}\\

\specialrule{\lightrulewidth}{0pt}{0pt}
\hiderowcolors
\multirow{2}{*}{\textbf{\methname}} & \textbf{73.60}\tiny{$\pm$0.3} & \textbf{77.77}\tiny{$\pm$0.2} & \textbf{68.30}\tiny{$\pm$0.6} & \textbf{72.70}\tiny{$\pm$0.2} & \textbf{52.19}\tiny{$\pm$0.6} & \textbf{61.63}\tiny{$\pm$0.1} & \textbf{48.64}\tiny{$\pm$0.0} & \textbf{59.23}\tiny{$\pm$0.4}\\
\addlinespace[-1.5px]
 & \gain{13.19} & \gain{4.68} & \gain{11.08} & \gain{5.06} & \gain{17.92} & \gain{11.57} & \gain{12.35} & \gain{9.42}\\

\bottomrule
\end{tabular}
}

\label{tab:results_ntu}
\end{table}
For the experimental comparison, we indicate with Joint Training (JT) the upper bound of our approach. It consists of training the model on the unified dataset (\ie, without splitting it into tasks). For the lower bound, we adopt an incremental training approach that does not employ tailored techniques against catastrophic forgetting. We refer to it as Fine Tuning (FT).

In~\cref{tab:results_ntu} we report the results for buffer sizes $\mathcal{M}_{size}$ of dimensions $500$ and $2000$. Following other works~\cite{buzzega2020dark,li2021else,boschini2022class}, we measure the recognition performance in terms of Final Average Accuracy (FAA), defined as:
\begin{equation}
    \operatorname{FAA} = \frac{1}{T} \sum_{i = 1}^{T} a_{\mathcal{T}_i},
\end{equation}
\noindent where $a_{\mathcal{T}_i}$ is the accuracy of the $i$-th task after the model has seen all $T$ of them. Additionally, we repeat each experiment three times, thus reporting the mean and standard deviation of the FAA.

As outlined by~\cref{tab:results_ntu}, the main competitor of this work, Else-Net~\cite{li2021else}, did not achieve performance comparable to those of the setting proposed by its authors, which devises a massive pre-training phase. Therefore, we can conclude that such a method suffers when trained from scratch.

Furthermore, even classical replay methods such as iCaRL~\cite{rebuffi2017icarl}, ER~\cite{riemer2018learning} and DER(\pp)~\cite{buzzega2020dark} outperform Else-Net. \methname reveals to be SOTA in the Class-IL skeleton-based action recognition domain, across both \ntu{60} and \ntu{120}. In particular, this holds when employing a \textit{masking ratio} of $30\%$; for higher percentages, we observe a decrease in performance, as discussed in the following. Significantly, the most substantial improvement is observed with a buffer size of 500 (surpassing the second-best, \ie, \derpp, when using a buffer size of 2000). This highlights the pivotal role of the sample quantity in the efficacy of replay methods. Consequently, it underscores the importance of researching techniques to increase sample numbers within a fixed buffer size.
\begin{figure}[t]
    \centering
    \includegraphics[width=\linewidth]{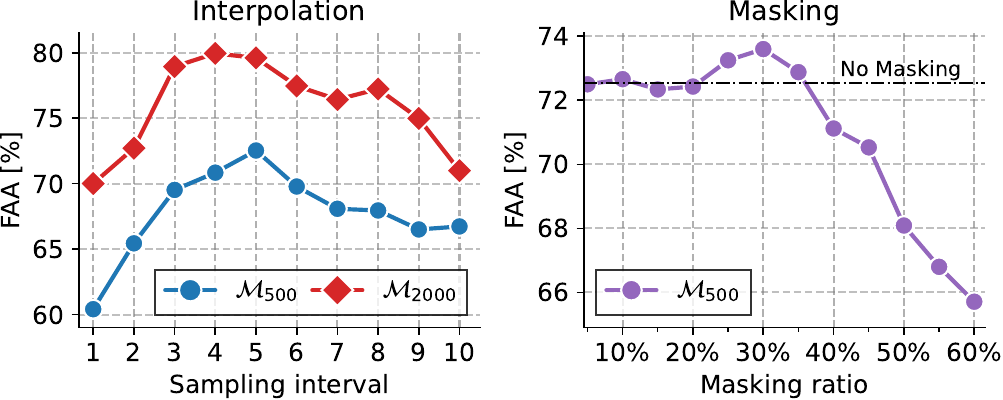}
    \caption{(\textit{left}) FAA for the \derpp baseline employing different values of the sampling interval $s$. (\textit{right}) FAA obtained by \methname as the masking ratio varies.}
    \label{fig:interpolator}
\end{figure}
\tit{On the sampling interval.} To further evaluate the effectiveness of our buffer strategy, we conduct a comparative study on varying \textit{sampling interval} $s$ (which we recall indicates the step length in the uniform sampling procedure). Given $s \in \mathbb{N}^\plus$, we obtain the \textit{compression ratio} as:
\begin{equation}
    \textit{compression ratio} = \frac{s - 1}{s} \cdot 100.
\end{equation}
We report in~\cref{fig:interpolator} (\textit{left}) the FAA at varying sampling interval $s$ for both the buffer sizes tested. For each tested sampling interval, we scale the buffer size accordingly (as documented in~\cref{subsec:method}). For instance, when $s=10$, a memory with a nominal capacity of $500$ examples could hence contain at most $s \cdot 500=5000$ (compressed) examples. As can be appreciated, the sampling interval $s=5$ (\ie, $80\%$ of \textit{compression}) yields the best results in terms of final accuracy. Namely, when sampling one skeletal pose every five frames, the memory buffer attains the best compromise between sample \textit{fidelity} (which can be achieved with lower sampling intervals) and sample \textit{diversity} (\ie, higher intervals). Moreover, we note that the presence of a prior compression phase ($s>1$) brings a stable and remarkable gain w.r.t. the standard replaying paradigm ($s=1 \rightarrow$ no compression at all). Such a result shows the crucial role of the trade-off between the quality and quantity of samples.

\tit{On the masking ratio.} We herein assess the impact of the \textit{masking ratio}, which indicates the number of frames discarded before feeding the input sequence to the model. The results are illustrated in~\cref{fig:interpolator} (\textit{right}) and reveal an increase in performance up to a value of $30\%$. For higher masking ratios, performance begins to decline, despite the notable efficiency gains (see~\cref{fig:efficie}). In quantitative terms, even with $50\%$ of masking, \methname achieves an acceptable final average accuracy of around $68\%$, while it decreases to $\approx 66\%$ with a masking ratio equal to $60\%$. Interestingly, both of these results are still higher than those of \derpp, the second-best method reported in~\cref{tab:results_ntu}.
\subsection{Ablations}
\label{subsec:ablations}
We herein report the ablative studies; all the experiments are performed on the XView modality of \ntu{60}.

\tit{On the importance of the reconstruction-based objective.} Our approach not only seeks good classification capabilities but also devises an auxiliary reconstruction term targeting the entire input sequence. To shed further light on the effects of such an auxiliary objective, we provide an ablative experiment in which we discard both the decoder module and the subsequent reconstruction loss. In doing so, we still apply random masking (testing two ratios equal to $30\%$ and $60\%$) and linear probing at the end of each task.

The results of these ablative studies are reported in~\cref{tab:recon_ablation}: remarkably, \methname experiences a significant performance drop when removing the decoder and the reconstruction loss, especially for the higher masking ratio of $60\%$. We consider such a finding as noteworthy, as it highlights the importance of auxiliary learning techniques when leveraging higher compression ratios to pursue efficiency.
\begin{table}[t]
\centering

\setlength{\tabcolsep}{7pt}
\renewcommand{\arraystretch}{1.2}

\begin{minipage}[b]{0.48\textwidth}
    \centering
    \caption{Impact of the reconstruction loss at different masking ratios.}
    \label{tab:recon_ablation}
    \begin{tabular}{lcc}
    \toprule
                    & \multicolumn{2}{c}{\textbf{Masking ratio}} \\
                    & 30\%           & 60\% \\ \midrule
    w/o recon. loss & 70.61          & 61.59 \\
    \rowcolor{lightgray}
    \methname       & \textbf{73.60} &  \textbf{65.72} \\

    \bottomrule
    \end{tabular}
\end{minipage}
\hfill
\begin{minipage}[b]{0.48\textwidth}
    \centering
    \caption{Ablative outcomes about sampling strategy and masking position.}
    \label{tab:mask_ablation}
    \begin{tabular}{ccc}
    \toprule
     & \multicolumn{2}{c}{\textbf{Position}} \\

    \textbf{Strategy} & \textit{pre} & \textit{post} \\
    \midrule
    \textit{Deterministic} & 72.08 &  72.43 \\
    \rowcolor{lightgray}
    \textit{Random} & 71.89 & \textbf{73.60} \\

    \bottomrule
    \end{tabular}
\end{minipage}\vspace{3pt}

\label{tab:results_ablations}
\end{table}
\tit{Masking strategy and positioning.} Our approach adopts a masking strategy that builds upon random guessing to drop frames, thus following most of the literature dealing with masked autoencoders. Herein, we want to compare our approach with a deterministic strategy, that drops one frame every $k$. We also assess different possible positions to introduce the masking operation. Specifically, \textit{post} indicates that masking is placed after splitting the sequence into tuples (see~\cref{subsec:preliminaries}), as carried out by our approach. Results for the combinations of these two alternatives are reported in~\cref{tab:mask_ablation}: as can be observed, the random strategy with post-hoc masking emerges as the best configuration.
\section{Conclusions}
Skeleton-based action recognition is a relevant task in modern human-centric Artificial Intelligence. We addressed such a long-standing computer vision task from the perspective of incremental learning, thus enabling those applications (\eg, sports analysis, rehabilitative healthcare) where the set of actions to be recognized may change over time. Differently from existing proposals dealing with action recognition, our work appoints \textit{efficiency} as a crucial aspect of an ideal incremental learner. 

Our method, named \methname, could be considered a step forward, as it achieves state-of-the-art performance with a remarkable reduction of the computational footprint (in terms of both memory and training time). In a few words, these capabilities derive from a proper application of input sub-sampling and random masking. Importantly, our experiments show that the addition of a reconstruction-based auxiliary objective grants further robustness in the presence of higher masking ratios, thus encompassing settings demanding efficiency. In future studies, we are going to deepen the concepts discussed in this paper, to apply our proposal even in the case of extreme masking (\eg, up to $95\%$).

\subsubsection{Acknowledgements} Andriy Sorokin was supported by Marie Sklodowska-Curie Action Horizon 2020 (Grant agreement No. 955778) for the project ``Personalized Robotics as Service Oriented Applications'' (``PERSEO''). Additionally, the research activities of Angelo Porrello have been partially supported by the Department of Engineering \quotationmarks{Enzo Ferrari} through the program FAR\_2023\_DIP -- CUP E93C23000280005.

%
%
%
\bibliographystyle{splncs04}
\bibliography{bibliography}
\newpage
\appendix
\renewcommand{\thetable}{\Alph{table}}
\renewcommand{\thefigure}{\Alph{figure}}
\renewcommand{\theequation}{\Alph{equation}}
\setcounter{figure}{0}    
\setcounter{table}{0}    
\setcounter{equation}{0}    
\pagenumbering{roman}
\section{\ntu{120}}
We hereby present \ntu{120}, the incremental version of the popular dataset introduced in~\cite{liu2019ntu}. Adhering to the structure delineated in~\cite{boschini2022class}, which entails $6$ tasks split into $10$ classes each, we expand upon it by incorporating additional $6$ tasks. This results in a total of $12$ tasks in an incremental scenario, each comprising a $10$-classes classification problem. Below, we present the order of these classes within each task.

\begin{enumerate}[start=1,label={\bfseries Task \arabic*}]
    \item \textit{put on glasses, cross hands in front, falling down, staggering, fan self, salute, throw, punch/slap, pushing, put palms together;}
    \item \textit{hand waving, sneeze/cough, type on a keyboard, nod head/bow, check time (from watch), brush teeth, hugging, wipe face, eat meal, shaking hands;}
    \item \textit{put on jacket, reading, take off a shoe, put on a shoe, tear up paper, reach into pocket, point to something, drink water, kicking something, sit down;}
    \item \textit{headache, pat on back, play with phone/tablet, writing, stand up, back pain, walking towards, shake head, walking apart, touch pocket;}
    \item \textit{take off glasses, point finger, brush hair, taking a selfie, giving object, take off jacket, take off a hat/cap, kicking, phone call, hopping;}
    \item \textit{rub two hands, nausea/vomiting, jump up, clapping, drop, chest pain, neck pain, put on a hat/cap, cheer up, pick up;}
    \item \textit{cutting nails, yawn, open a box, flick hair, cutting paper, hit with object, hush, knock over, make victory sign, put on bag;}
    \item \textit{toss a coin, follow, cross arms, cross toe touch, capitulate, shake fist, support somebody, move heavy objects, whisper, put on headphone;}
    \item \textit{take a photo, tennis bat swing, shoot with gun, sniff/smell, thumb down, butt kicks, take object out of bag, counting money, grab stuff, play magic cube;}
    \item \textit{snap fingers, shoot at basket, blow nose, cheers and drink, side kick, exchange things, apply cream on face, step on foot, bounce ball, throw up cap/hat;}
    \item \textit{high-five, take off headphone, fold paper, arm swings, rock-paper-scissors, make OK sign, squat down, carry object, stretch oneself, thumb up;}
    \item \textit{wield knife, staple book, arm circles, put object into bag, run on the spot, ball up paper, juggle table tennis ball, open bottle, apply cream on hand, take off bag.}
\end{enumerate}
\begin{table}[t]
\centering

\rowcolors{2}{lightgray}{}
\setlength{\tabcolsep}{6pt}
\renewcommand{\arraystretch}{1.2}

\caption[sklMAE]{The skeleton-based masked autoencoder architecture adopted.}

\resizebox{0.95\linewidth}{!}{\begin{tabular}{ccc}
\toprule
\multicolumn{3}{c}{\textbf{sklMAE} - Input Size: $3 \times 120 \times 25$} \\

\midrule

\textbf{Layer Name} & \textbf{Output Size} & \textbf{Layer Details} \\

\specialrule{\lightrulewidth}{0pt}{0pt}

\texttt{division_in_tuples} & $3 \times 20 \times 150$ & \Gape[0pt][2pt]{\makecell{\texttt{reshape(shape[0], } \\ \texttt{shape[1] // 6, shape[2] * 6)}}} \\
\texttt{input_map} & $32 \times 20 \times 150$ & \Gape[0pt][2pt]{\makecell{Conv2D $1 \times 1$, 32 \\ BatchNorm2D}} \\
\texttt{masking} & $32 \times ((1 - \eta) \cdot 20) \times 150$ & \Gape[0pt][2pt]{\makecell{Random masking, \\ $\eta \in [0, 1)$}} \\
\texttt{encoder_1} & $32 \times ((1 - \eta) \cdot 20) \times 150$ & \Gape[0pt][2pt]{\makecell{STTFormer layer, \\ $\texttt{qkv_dim} = 16$}} \\
\texttt{encoder_2} & $32 \times ((1 - \eta) \cdot 20) \times 150$ & \Gape[0pt][2pt]{\makecell{STTFormer layer, \\ $\texttt{qkv_dim} = 16$}} \\
\texttt{encoder_3} & $64 \times ((1 - \eta) \cdot 20) \times 150$ & \Gape[0pt][2pt]{\makecell{STTFormer layer, \\ $\texttt{qkv_dim} = 32$}} \\
\texttt{encoder_4} & $64 \times ((1 - \eta) \cdot 20) \times 150$ & \Gape[0pt][2pt]{\makecell{STTFormer layer, \\ $\texttt{qkv_dim} = 32$}} \\
\texttt{encoder_5} & $128 \times ((1 - \eta) \cdot 20) \times 150$ & \Gape[0pt][2pt]{\makecell{STTFormer layer, \\ $\texttt{qkv_dim} = 64$}} \\
\texttt{encoder_6} & $128 \times ((1 - \eta) \cdot 20) \times 150$ & \Gape[0pt][2pt]{\makecell{STTFormer layer, \\ $\texttt{qkv_dim} = 64$}} \\
\texttt{encoder_7} & $128 \times ((1 - \eta) \cdot 20) \times 150$ & \Gape[0pt][2pt]{\makecell{STTFormer layer, \\ $\texttt{qkv_dim} = 64$}} \\
\texttt{encoder_8} & $128 \times ((1 - \eta) \cdot 20) \times 150$ & \Gape[0pt][2pt]{\makecell{STTFormer layer, \\ $\texttt{qkv_dim} = 64$}} \\

\specialrule{\lightrulewidth}{0pt}{0pt}

\texttt{classifier} & $128  \times 60$ & \Gape[0pt][2pt]{\makecell{Average pooling \\ Linear layer}} \\

\specialrule{\lightrulewidth}{0pt}{0pt}

\texttt{mask_concat} & $128 \times 20 \times 150$ & \Gape[0pt][2pt]{\makecell{Concatenation of the \\ \texttt{[MASK]} tokens}} \\
\texttt{decoder_1} & $64 \times 20 \times 150$ & \Gape[0pt][2pt]{\makecell{STTFormer layer, \\ $\texttt{qkv_dim} = 64$}} \\
\texttt{decoder_2} & $32 \times 20 \times 150$ & \Gape[0pt][2pt]{\makecell{STTFormer layer, \\ $\texttt{qkv_dim} = 32$}} \\
\texttt{decoder_3} & $32 \times 20 \times 150$ & \Gape[0pt][2pt]{\makecell{STTFormer layer, \\ $\texttt{qkv_dim} = 16$}} \\
\texttt{output_map} & $3 \times 20 \times 150$ & \Gape[0pt][2pt]{Conv2D $1 \times 1$, 3} \\

\bottomrule
\end{tabular}
}

\label{tab:output_dims}
\end{table}
\section{Skeleton-based MAE}
In~\cref{tab:output_dims}, we present the output dimensions and layer specifications of our Masked AutoEncoder-inspired skeleton-based architecture. As outlined in the paper, we utilize a modified version of STTFormer~\cite{qiu2022spatio} for the backbone, which is reduced in width. The \texttt{qkv_dim} parameter determines the embedding dimensions of the queries, keys, and values, respectively. The number of attention heads in each layer is consistently set at $3$. Regarding the IFFA modules, which incorporate temporal pooling aggregation, we opt for a default kernel size of $3 \times 1$ and a padding of $(1, 0)$. Lastly, the linear classifier is intended for the 60 classes of \ntu{60}.
\begin{figure}[t]
    \centering
    \includegraphics[width=0.7\textwidth]{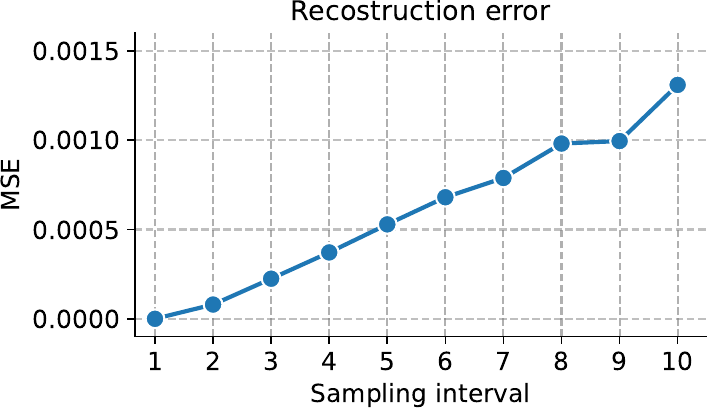}
    \caption{Figure showing the reconstruction error of the interpolation procedure as the sampling interval varies.}
    \label{fig:reconstruction}
\end{figure}
\section{Interpolation reconstruction error}
We display in~\cref{fig:reconstruction} the reconstruction error of the interpolation procedure as the sampling interval varies. To quantify the distance between the reconstructed and the ground truth samples, we employ a simple yet effective Mean Squared Error (MSE) metric.

\end{document}